\documentclass[10pt,twocolumn,letterpaper]{article}

\usepackage{cvpr}
\usepackage{times}
\usepackage{epsfig}
\usepackage{graphicx}
\usepackage{amsmath}
\usepackage{amssymb}
\usepackage{subfigure}
\usepackage{lipsum}
\usepackage[pagebackref=true,breaklinks=true,letterpaper=true,colorlinks,bookmarks=false]{hyperref}

\cvprfinalcopy % *** Uncomment this line for the final submission

\def\cvprPaperID{3213} % *** Enter the CVPR Paper ID here

\usepackage[usenames,dvipsnames,svgnames,table]{xcolor}

\ifcvprfinal\fi
\begin{document}

%%%%%%%%% TITLE
\title{Neuron-level Selective Context Aggregation for Scene Segmentation}

\author{Zhenhua Wang*\\
Hebrew University of Jerusalem\\
{\tt\small zhenhua.wang@mail.huji.ac.il}
\and
Fanglin Gu*\\
Shandong University\\
{\tt\small gfl699468@mail.sdu.edu.cn}
\and
Dani Lischinski\\
Hebrew University of Jerusalem\\
{\tt\small danix@mail.huji.ac.il}
\and
Daniel Cohen-Or\\
Tel Aviv University\\
{\tt\small dcor@tau.ac.il}
\and
Changhe Tu\\
Shandong University\\
{\tt\small chtu@sdu.edu.cn}
\and
Baoquan Chen\\
Shandong University\\
{\tt\small baoquan@sdu.edu.cn}
}

\maketitle
\newcommand\blfootnote[1]{%
\begingroup
\renewcommand\thefootnote{}\footnote{#1}%
\addtocounter{footnote}{-1}%
\endgroup
}
\blfootnote{* equal contribution}
%\thispagestyle{empty}

%%%%%%%%% ABSTRACT
\begin{abstract}
   Contextual information provides important cues for disambiguating visually similar pixels in scene segmentation. In this paper, we introduce a neuron-level Selective Context Aggregation (SCA) module for scene segmentation, comprised of a contextual dependency predictor and a context aggregation operator. The dependency predictor is implicitly trained to infer contextual dependencies between different image regions. The context aggregation operator augments local representations with global context, which is aggregated selectively at each neuron according to its on-the-fly predicted dependencies.
   The proposed mechanism enables data-driven inference of contextual dependencies, and facilitates context-aware feature learning. The proposed method improves strong baselines built upon VGG16 on challenging scene segmentation datasets, which demonstrates its effectiveness in modeling context information.
\end{abstract}

%%%%%%%%% BODY TEXT
\section{Introduction}

\begin{figure*}[htb]
\centering
\includegraphics[width=0.98\textwidth]{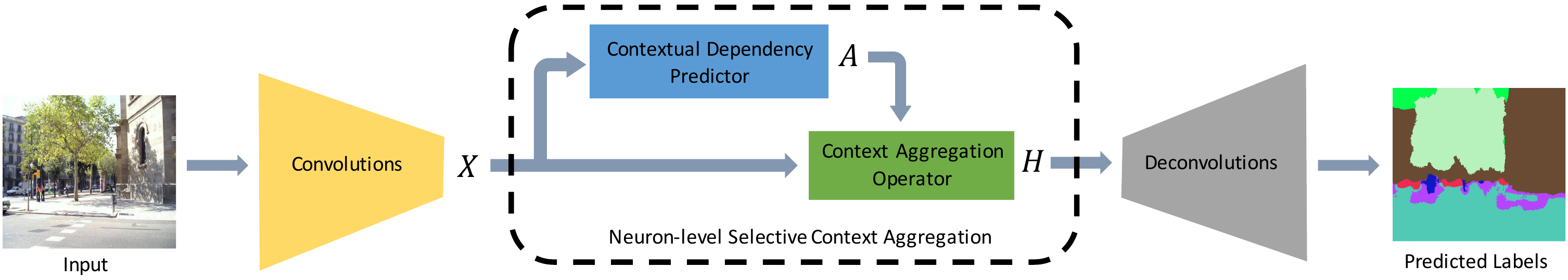}
\caption{An overview of the proposed scene segmentation network architecture.}
\label{fig_workflow}
\end{figure*}

Scene segmentation is a long standing fundamental problem in computer vision, which aims to associate a semantic object category label with each pixel in an image of a scene.
It has long been recognized that taking into account the semantic context between image regions can significantly improve the performance of scene segmentation algorithms~\cite{Krahenbuhl_11,Mottaghi_14,Shotton_09,Yang_14}.

As is the case in many other areas of computer vision, the state-of-the-art in scene segmentation has been transformed through the use of deep neural networks.
Long \etal~\cite{Long_15} proposed a fully convolutional network which can be trained end-to-end to predict a dense labeling map, capable of extracting coarse object shapes. This approach was extended by Noh \etal~\cite{Noh_15} by introducing multiple deconvolutional layers to gradually capture the finer details.
In these pure CNN-based methods, the representations of neurons in the convolutional feature maps are extracted from limited receptive fields, capturing only local information.

To address this problem, Liu \etal \cite{Liu_15} propose to use the average feature of a layer as global context to augment the local features at each neuron. Yu and Koltun \cite{Yu_2016} introduce dilated convolutions to systematically aggregate multi-scale contextual information without losing resolution.
Zhao \etal \cite{Zhao_17} exploit global context information by different region-based context aggregation through
pyramid pooling.
Although these methods improve the segmentation accuracy, they aggregate context in a predefined manner, which does not depend on the input image. Thus, they lack internal mechanisms for handling non-uniform dependencies among different image regions.

In another line of research, Shuai \etal~\cite{Shuai_17} proposed Directed Acyclic Graph Recurrent Neural Networks (DAG-RNNs) to leverage contextual information, thereby enhancing the representation ability of convolutional features.
Their approach has demonstrated significant advances over the state-of-the-art on three challenging datasets.
However, due to the gradient vanishing problem of RNNs~\cite{Bengio_94, Pascanu_13}, the contextual information can only be propagated to nearby neurons, thus the long-range dependencies cannot be well captured.
In fact, Shuai \etal~\cite{Shuai_16} report that using DAG-RNNs induced from 8-connected grids, significantly outperforms the 4-connected case. Moreover, due to the usage of RNNs, the hidden vectors of DAG-RNNs must be sequentially updated, making it less efficient than pure CNN-based methods.

In this work, we propose a new approach for selectively aggregating contextual information at each neuron, and doing so in an input-dependent manner, by introducing a novel context aggregation module.
This module takes as input the feature map of a convolutional layer, and outputs a context-aware feature map.
Specifically, this module consists of two main components: a \emph{contextual dependency predictor} and a \emph{context aggregation operator}.
The dependency predictor is an auxiliary network that infers a coefficient representing the degree of dependency between each pair of neurons. Note that these coefficients are input-dependent. The context aggregator densely connects each output neuron to all the input neurons according to its predicted coefficients, yielding feature maps with enhanced context representation ability. These feature maps are then fed to a deconvolutional network to predict a semantic label map (see Fig.~\ref{fig_workflow}).
We refer to the proposed module as a neuron-level Selective Aggregation (SCA) module, since each of its output neurons uses a different set of weights to aggregate contextual features from the input neurons.
%attend to different neurons to engage context information.
%\danix{See my previous comment on using the term attention... we should find something better.}

In summary, our work makes three main contributions:
\begin{itemize}
\item We present a novel context aggregation module which selectively injects global context information into local representations, based on image-specific dependencies among neurons in the deep convolutional feature map.
\item We  introduce a contextual dependency predictor that learns to infer image-specific dependencies between neurons. The resulting dependency coefficients allow each neuron to selectively aggregate context information from the entire image.
\item We apply the proposed network on the scene segmentation task, and demonstrate its effectiveness and potential on two large datasets.
\end{itemize}

The rest of the paper is organized as follows: Sec.~\ref{sec_rela} discusses some work related to our method, Sec.~\ref{sec_approach}
introduces the two components of the proposed new module, Sec.~\ref{sec_detail} gives the implementation details, and, finally, Sec.~\ref{sec_exp} reports the results of our experiments.

\section{Related Works\label{sec_rela}}

The semantic segmentation problem has been addressed by a wide variety of methods in recent years.
Since visually similar pixels are locally indistinguishable, 
a major question that arises is how to leverage contextual information for disambiguation.
Many methods rely on Probabilistic Graphical Models (PGMs), \eg Markov Random Fields (MRFs) and Conditional Random Fields (CRFs), to account for context~\cite{Shotton_09, Krahenbuhl_11, Yao_12, Zhang_12, Roy_14}.
These methods usually require a pre-segmentation, such as superpixels, from which to extract features, and they are
usually inefficient for inference due to the iterative solving of local beliefs.

Due to the recent advances in deep neural networks, 
the mainstream methods of  semantic segmentation usually adopt Convolutional Neural Networks (CNNs) as their basic
component.
Farabet~\etal~\cite{Farabet_13} feed their CNN with a multi-scale pyramid of images, covering
a large context.
Pinheiro~\etal~\cite{Pinheiro_14} encode long range pixel label dependencies by using a recurrent CNN architecture.
Sharma~\etal~\cite{Sharma_14} use a recursive neural network to recursively aggregate contextual information from local neighborhoods to the entire image and then propagate the global context back to individual local features.
Shuai~\etal~\cite{Shuai_15} use a non-parametric model to represent global context, and transfer 
such information to local features extracted by CNN.
Long~\etal~\cite{Long_15} propose a fully convolutional network to directly predict label map from the input image. Noh~\etal~\cite{Noh_15} extend~\cite{Long_15} by introducing multi-layer deconvolution networks.

Liu \etal \cite{Liu_15} propose to use the average feature of a layer as global context to augment the features at each neuron. Yu and Koltun \cite{Yu_2016} introduce dilated convolutions to systematically aggregate multi-scale contextual information without losing resolution.
Zhao \etal \cite{Zhao_17} exploit global context information by different region-based context aggregation through
pyramid pooling. 
Shuai \etal \cite{Shuai_17} utilize DAG-RNNs to embed context into local features.
Different from all these methods, which use an input-independent and fixed manner to aggregate context, our method introduces a dynamic mechanism which aggregates context selectively at each neuron, and does so in an input-dependent manner. 
Our approach of using an auxiliary network to achieve input-dependent processing is similar in spirit to spatial transformer networks~\cite{Jaderberg_15} and deformable convolution networks~\cite{Dai_17}

%\zhwang{TODO: maybe add more related works about attention if we do not change the term "Attentional"}

\section{Our Approach\label{sec_approach}}

%\danix{Rather than beginning by stating what are the components in the order in which they appear in the pipeline, a higher-level overview is needed. It should motivate what the attention predictor AP and the context aggregator CA are attempting to do, and how they achieve it. The fact that the weights produced by the AP are input-dependent and not fixed for different images should be emphasized. This seems to be a major novel aspect compared to the fixed architecture such as the DAG-RNN, or dilated convolutions. ZH, I need you to try and write the first draft of this overview, then I can try to refine it.}~\zhwang{please check the next paragraph. But should we move it to introduction? I said something similar in the introduction.}

Contextual information has been widely used in computer vision applications~\cite{Krahenbuhl_11,Mottaghi_14,Shotton_09,Yang_14}.
In general, context can refer to any global information that has a facilitative effect on representation ability.
Note, however, that a local representation may be influenced differently by different semantic categories of pixels in its context.
For example, the presence of `sea' pixels in an image is a more important clue for classifying nearby sand-colored pixels as `beach', compared to `mountain' pixels, since `sea' and `beach' have higher co-occurrence correlation.
However, previous methods~\cite{Liu_15,Yu_2016,Zhao_17} usually embed global context into a local representation in a fixed and uniform way, thereby lacking internal mechanisms to effectively account for non-uniform dependencies among different image regions.

To enhance the ability of handling non-uniform contextual information, we propose an input-dependent way to selectively incorporate context into the local representation.
Fig.~\ref{fig_workflow} shows an overview of our proposed segmentation network architecture.
Given an input image $\mathbf{I}$, the convolution layers gradually extract its abstract features, yielding a high-level feature map $\mathbf{X}$. Next, a neuron-level Selective Context Aggregation (SCA) module is applied onto $\mathbf{X}$ to augment it with context information.
The SCA module consists of two components.
The first component is the contextual dependency predictor network, which takes $\mathbf{X}$ as input, and through several hidden layers predicts a matrix $\mathbf{A}$, which consists of the \emph{dependency coefficients} between all pairs of neurons in $\mathbf{X}$.
A dependency coefficient $a_{ij}$ attempts to predict the extent to which neuron $\textbf{x}_j$ should be accounted for in the context of neuron $\textbf{x}_i$.
%decide the regions of the input feature map to pay attention to when engaging context.
%\danix{The term "dependency coefficients" is still a bit too general, not clear exactly what is considered a dependency. Can we be a bit more specific/explicit, and more precise, here?}
%\zhwang{I think no one knows the what the exact depency it is. The predicotr is implicitly supervised by the final segmentation loss. It will be something that can help to minimize the segmentation loss, but with no intuitive meanning. This is why I firstly call it "abstract correlation".  It is all determined by the learning process, and not necessary to coincide with human intuition}
Next, the context aggregation operator uses the predicted dependency matrix $\mathbf{A}$ to transform the input feature map $\mathbf{X}$ into a context-aware feature map $\mathbf{H}$.
Finally, several deconvolutional layers are applied on $\mathbf{H}$ to predict the final semantic segmentation label map.

%\danix{What is the reason to describe below the aggregation first and the predictor second? Would it make more sense to reverse the order?}
%\zhwang{In order of computation, it is first predictor then aggregator, as i wrote in the first paragraph of sec. 3 Latter, the reason I first introdcue aggregator that is that I want write it more general. The aggregator is a general operator which selectively aggregates context according to $a_ij$. However, in principle, $a_ij$ is not necessary to be predicted by an auxilarity network. it can be manually set according to prior knowlege (e.g., all one matrix, dianal one matrix, or co-occurrence matrix that obtained from statistics outside the network). Only if we want to learn the dependency jointly, we need to plug a predictor network. Since we adopt the data-driven approach to learn dependency, I then introduce how I design the predictor network to fullfil our requirements. If I firstly introduce the predictor, it sounds like the aggregator must rely on the output of the predictor. But in principle, it is not true. How do you think?}

\subsection{Context Aggregation Operator}
\begin{figure}[tb]
\centering
\includegraphics[width=0.45\textwidth]{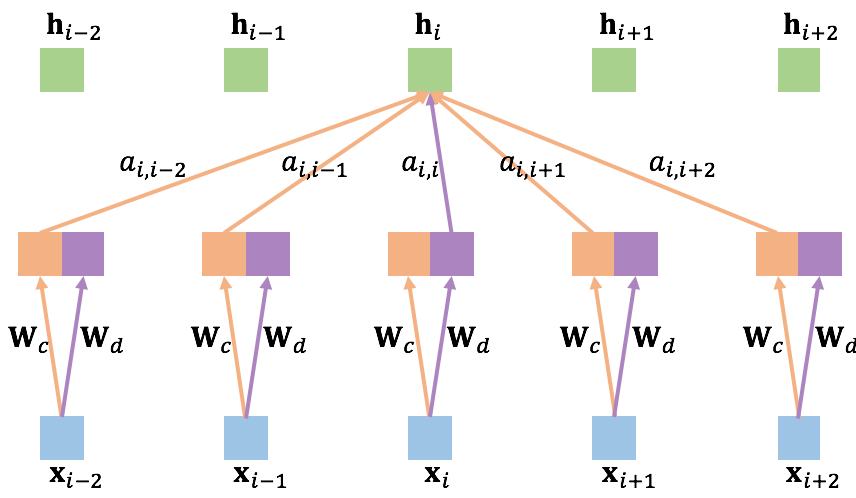}
\caption{An illustration of the proposed context aggregation operator on a 1D feature map.}
\label{fig_CA}
\end{figure}

%\danix{Add 1-2 sentences that describe the context aggregator in simple informal language before delving into the precise formal equations. Hopefully, the motivation has been made clear above already, but the implementation should be presented in a top-down fashion, and not bottom-up.}~\zhwang{addressed. see below texts}

Generally, a context aggregation operator aims to encode global contextual information into a local neuron representation. It takes as input a feature map $\mathbf{X}$, and produces a context-aware feature map $\mathbf{H}$ with the same spatial dimensions (but possibly with a different number of channels).
To fully exploit the context, the receptive field of this operator should cover the full size image.
A naive approach is to use a fully-connected layer in which each output neuron is connected to all the input neurons. To be more specific, let $\mathbf{x}_i \in \mathbb{R}^N$ and $\mathbf{h}_i \in \mathbb{R}^M$ denote the feature vectors of the $i$-th neuron in $\mathbf{X}$ and $\mathbf{H}$, respectively.
Using a fully-connected layer each neuron in $\mathbf{H}$ is given by:
\begin{equation}
\mathbf{h}_i = \sum_{j=1}^{n}{\mathbf{W}_{ij}\mathbf{x}_j}
\label{eq_context}
\end{equation}
where $n$ is the number of input neurons in $\mathbf{X}$, and $\mathbf{W}_{ij} \in \mathbb{R}^{M\times N}$ is the weight matrix.

As can be seen, $\mathbf{W}_{ij}$ is particular to the $i$-th input neuron and the $j$-th output neuron. Thus, the learned weights are location sensitive, which is undesirable for dense label prediction, as the same category of pixels can appear anywhere in the image.
Furthermore, after training, $\mathbf{W}_{ij}$ is fixed during test time, and thus the context aggregation is input-independent.
Finally, the fully-connected layer consumes a huge amount of learnable parameters, making it easy to overfit the training data.

To address all these problems, we introduce a new aggregation operator.
Our operator first uses two $1\times 1$ convolutions to extract two different features, the \emph{identity feature} and the \emph{context feature}, for each neuron. The intuition is that the feature used to describe the neuron itself and the feature that describes its contribution to the context of other neurons may be different.
Then, the output feature of a neuron is produced by linearly combining its own identity feature with the context features of all the other neurons, according to some input-dependent coefficients.
Formally, the computation of the output neuron $\mathbf{h}_i$ is defined as:
\begin{equation}
\mathbf{h}_i = a_{ii}\mathbf{W}_d\mathbf{x}_i+\frac{1}{\sum_{j=1,j\neq i}^{n}a_{ij}}\sum_{j=1,j\neq i}^{n}{a_{ij}\mathbf{W}_c\mathbf{x}_j}
\label{eq_context}
\end{equation}
where $a_{ij}$ is a coefficient which controls the amount of information taken from $\mathbf{x}_j$ to $\mathbf{h}_i$, $\mathbf{W}_d \in \mathbb{R}^{M\times N}$ and $\mathbf{W}_c \in \mathbb{R}^{M\times N}$
are the weight matrices for extracting the identity and context features, respectively.
In practice, we always set $a_{ii}$ to 1 so that the output neuron takes all of its identity feature.
Fig.~\ref{fig_CA} illustrates the computation procedure of the proposed operator for a single output neuron on a 1D feature map.
%\danix{Also, is there any activation function involved as well, or is it really just a linear combination without any non-linearities at the end?} \zhwang{The proposed layer is a linear operator, like convolution layer and fc-layer. The activation function can be attached on top of the proposed operate. They are separate.}

Notably, this new operator has several appealing properties.
First of all, the matrices $\mathbf{W}_d,\mathbf{W}_c$ are shared across neurons, reducing the parameter space from $n^2NM$ (for a fully-connected layer) to $2NM$.
Secondly, the operation preserves spatial information, thus it is suitable to be applied on dense prediction tasks.
Thirdly, the coefficients $a_{ij}$ are not fixed parameters, and are capable of representing arbitrary dependencies between different pairs of neurons. These dependencies may be manually defined, according to some prior knowledge, or automatically inferred for the specific input $\mathbf{X}$.
Here we adopt a data-driven paradigm, and introduce an auxiliary contextual dependency predictor (Sec.~\ref{subsec_ap}) that learns to infer the dependencies among neurons and provide on-the-fly input-dependent predictions for the proposed context aggregation operator.

%\danix{In what sense are they implicit?}~\zhwang{In the sense that there is no direct/explicit supervision to learn the dependencies. The attention predictor is jointly trained with the main segmentation network. It learns to how to predict dependencies for different pairs of neurons, to help minimize the overall segmentation costs during training. It is implicitly supervised by the segmentation cost. }\danix{I see. We should make sure that we say this somewhere in the text. Probably in the overview of this section or in the subsection describing the predictor.}

To allow backpropagation through the proposed operator, we must define the gradients of the loss $\mathcal{L}$ with respect to both the input feature map neurons $\mathbf{x}_i$ and the coefficients $a_{ij}$.
These gradients are given by
\begin{equation}
\label{eq_bp_x}
\frac{\partial \mathcal{L}}{\partial \mathbf{x}_i} =
a_{ii}\mathbf{W}_d^{\mathrm{T}}\frac{\partial \mathcal{L}}{\partial \mathbf{h}_i}+\sum_{k\neq i}^n{
\frac{a_{ki}}{\sum_{j\neq k}^n{a_{kj}}}\mathbf{W}_c^{\mathrm{T}}\frac{\partial \mathcal{L}}{\partial \mathbf{h}_k}},\forall i\in[1,n]
\end{equation}
and
\begin{equation}
\label{eq_bp_a}
\frac{\partial \mathcal{L}}{\partial a_{ij}} = \sum
 \frac{\partial \mathcal{L}}{\partial \mathbf{h}_i} \odot \frac{ \sum_{k\neq i}^{n}a_{ik} (\mathbf{W}_c\mathbf{x}_j -\mathbf{W}_c\mathbf{x}_k)}{(\sum_{k\neq i}^{n}a_{ik})^2}, \forall j\neq i
\end{equation}
where $\frac{\mathcal{L}}{\partial \mathbf{h}_i}$ denotes the loss gradients with respect to $\mathbf{h}_i$ received from top layer, $\odot$ denotes the element-wise multiplication, and
the outermost sum in \eqref{eq_bp_a} is over the feature channel.

\subsection{Contextual Dependency Predictor}
\label{subsec_ap}

\begin{figure*}[htb]
\centering
\includegraphics[width=0.9\textwidth]{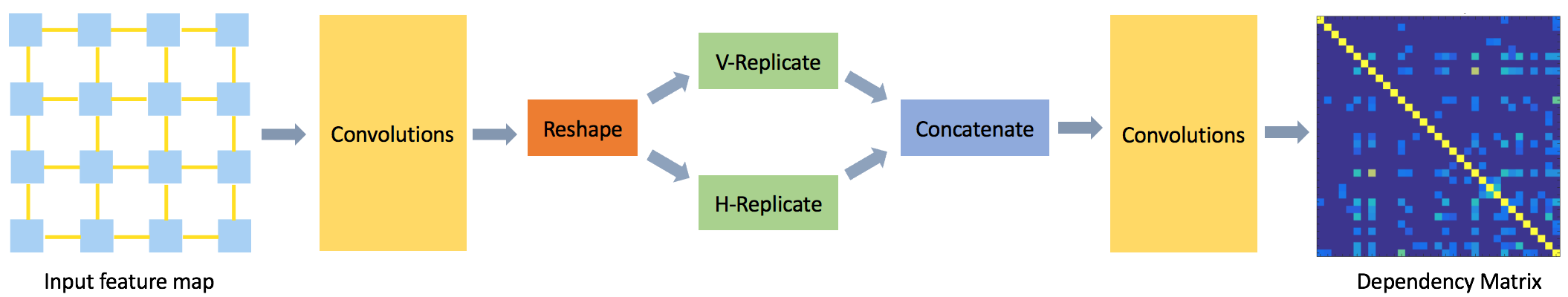}
\caption{The architecture of the contextual dependency predictor. V-Replicate denotes a layer which vertically replicates the input 1D feature map to a 2D square feature map, and H-Replicate denotes the horizontally replicated one.}
\label{fig_APN}
\end{figure*}

%\danix{The exposition below is now a bit more clear to me, but still I am bothered that we keep talking about dependencies among neurons, without ever explaining what these dependencies are and without showing any concrete examples, that would make the reader understand what kind of dependencies this network learns to predict.}
%\zhwang{As we discussed in email, I will show some visual examples.}
%\dc{By dependency I think he simply means a strong correlation between their values. What I don't understand is why we need to help the network to learn... if neuron i and j are likely to be similar, why should I hint that to the network, let it learn it itself...}
%\zhwang{We need a mechanism to enable each neuron to select different context. Otherwise,each neuron will uniformly pick the context, like the previous method who simply take the averaged feature across all spatial dimension as global context.}

The contextual dependency predictor (CDP) aims to estimate the dependencies between neurons of the input features. More precisely, given a pair of neurons $(\mathbf{x}_i, \mathbf{x}_j)$, the goal is to predict the weight that the context feature extracted from $\mathbf{x}_j$ should be given when aggregating the context of neuron $\mathbf{x}_i$. Since each pair may be assigned a different weight, the global context information is encoded selectively into each neuron's local representation.

The dependency function between neurons may be modeled by a neural network.
Typically, Siamese networks, which are popular for tasks that involve finding inherent relationships between two inputs, can be used to learn such a function.
Training Siamese networks requires paired inputs and their corresponding ground truths.
However, our input feature map is not organized as neuron pairs, and, more importantly, we do not have well-defined ground truth for the dependencies among such pairs.

Our approach is thus to learn the dependency function implicitly, by training the CDP jointly with the context aggregation operator (as well as the rest of the network) to minimize the final segmentation loss.
The CDP should take a feature map $\mathbf{X}$ with $n$ neurons as input, and output an $n\times n$ matrix $\mathbf{A}$, in which each element $a_{ij}$ represents the dependency coefficient between neuron $\mathbf{x}_i$ and $\mathbf{x}_j$, for the context aggregation operator.

To achieve this goal, we propose a new network architecture described below; also see Figure \ref{fig_APN}.
Firstly, several $1\times 1$ convolutional layers are applied on $\mathbf{X}$ to map the representation of each neuron into a higher-level feature space.
Intuitively, the purpose of this mapping is to prevent the network from simply computing linear correlations between the neurons; instead, the network is able to explore higher-level semantic relationships that may prove useful for the final semantic segmentation.

In order to efficiently predict dependencies for all neuron pairs, we
use a simple trick to construct a large feature map in which each location stores a concatenation of the mapped features of a particular neuron pair.
Concretely, we reshape the 2D feature map to a 1D feature vector, expand it to two 2D square feature maps using horizontal and vertical replication, respectively, and then concatenate these two replicated feature maps through the depth channel.
Using the resulting large feature map, we simply apply one more $1\times 1$ convolutional layer to produce the dependency matrix.
%An overview of the architecture is shown in Fig.~\ref{fig_APN}.

According to Eq.~\eqref{eq_bp_a}, the CDP can receive gradients from upper layers which enable it to be jointly trained with the main network without using extra supervision.
It learns how to selectively incorporate context information, by producing appropriated dependency coefficients for different pairs of neurons, to help minimize the overall segmentation cost during training.

%\dc{This network that computes a dependency matrix using the loss of segmentation, is a strong contribution that has a wider potential. As you said, it is an alternative for a siamese for small set, maybe. I feel we can sell it better, not just as a little solution for our needs here. Maybe we need to pose the question how to learn such dependencies where we have no GT... then say that we can use an implicit loss of some other task }
%\zhwang{Yes, I agree in general it has more potential. But the inspiration comes from spatial transformer network where they implicitly learn the geometric transformation. In this sense, we are similar in general.}

\section{Implementation Details\label{sec_detail}}
\subsection{Network Architecture}

The convolution layers of our network are adopted from the ImageNet pre-trained VGG16 network, i.e., all the layers before the $5$-th pooling layer. 
Following~\cite{Yu_2016}, we remove the last two pooling and striding layers, and replace all subsequent convolutions with dilated convolutions
of appropriate factors to make the last convolution layer 
produce a feature map that is $4$ times larger.

For the proposed SCA module, the dimensions of $\mathbf{W}_c$ and $\mathbf{W}_d$ in the context aggregation operator are both set to be $512\times 512$. The contextual dependency predictor uses three $1\times 1\times 128$ convolution layers to extract features, and uses a $1\times 1 \times 1$ convolution layer on the concatenated features to predict dependent coefficients.

The deconvolution module is similar to that of a FCN~\cite{Long_15}. Concretely, we convert the first fully-connected layer of VGG16 to a $7\times 7 \times 4096$ convolution layer, and convert the second one to a $1\times 1\times 4096$ convolution layer. Then, a $1\times 1\times \#\mathit{classes}$ convolution is applied to predict scores for all classes in the dataset at each location.
Finally, a bilinear upsampling layer is applied to resize the coarse score map to have the same resolution as the input.

\subsection{Loss Re-weighting}
The class distributions in scene labeling datasets are usually highly unbalanced, thus, it is common to increase the weights of rare classes during training to improve their accuracies~\cite{Farabet_13, Shuai_16}. In our work, we simply adopt the re-weighting strategy proposed in~\cite{Shuai_16}. Specifically, the weight for class $c_i$ is defined as $w_i = 2^{\lceil log_{10}(\eta / f_i)  \rceil }$ where $f_i$ is the frequency of $c_i$ and $\eta$ is a dataset dependent scalar defined by the $85\%/15\%$ rule on frequent$/$rare classes.

\subsection{Training and Testing}

For training, since every component of the proposed segmentation network is differentiable, errors can be backpropagated to all network layers and parameters, making the network trainable using any gradient-based optimization method. In practice, we use min-batch gradient descent with momentum. 
During training, we use the ``poly'' learning rate policy, where current learning rate
equals to the base learning rate multiplied by $(1-\frac{iter}{max iter})^{0.9}$. The base learning rate for the bottom convolution module is set to $10^{-3}$,
while it is set to $10^{-2}$ for both the SCA and the deconvolution modules.
Due to the GPU memory limitations, we set the batch size to 3. 
The number of training epochs is set to 50.
To augment training data, similarly to most of the previous methods, we adopt random horizontal flipping for all datasets.
At test time, we simply do a forward pass of the test image to obtain the estimated label map.

\section{Experiments\label{sec_exp}}

Similarly to most previous works, we report results on three different evaluation metrics: Per-Pixel Accuracy (PPA), Class Average Accuracy (CAA) and mean Intersection over Union (mIoU).
We extensively compare our proposed method with the-state-of-the-art methods on two large scene segmentation datasets:

\paragraph{PASCAL Context}~\cite{Mottaghi_14} dataset contains $10103$ images with an approximate resolution of
$375\times 500$, in which $4999/5104$ images are used for training/testing.
These images originally come from Pascal VOC 2010 dataset, and are augmented with pixel-wise annotations of $540$ classes.
Following~\cite{Mottaghi_14}, we only consider the most frequent $59$ classes in the evaluation. To enable batch-based training, we rescale and pad each image to be $448\times448$.

\paragraph{COCO Stuff}~\cite{Caesar_16} dataset contains $10000$ images in various resolutions, in which $9000/1000$ images are used for training/testing.
These images originally come from the Microsoft COCO dataset~\cite{Lin_14} with pixel-wise annotations of $80$ \textit{Thing} categories. They are further augmented with $91$ \textit{Stuff} categories. In total, there are $171$ categories in this dataset.

\subsection{Hyper-parameters Study}
To be fair, we conduct the hyper-parameter study on a third dataset, SIFT Flow~\cite{Liu_09}. This dataset is smaller compared to PASCAL \textit{Context} and COCO \textit{Stuff}, making it faster for tuning. It contains $2488/200$ images for training/testing with a resolution of $256\times 256$, and are annotated with $33$ classes.

Specifically, we study two hyper-parameters of the attention predictor:
the number $K_l$ of convolution layers for feature extraction and
the number $K_f$ of output feature channels of convolution in each layer.
The results are shown in Tab.~\ref{tb_layer} and Tab.~\ref{tb_chn}, respectively.
We note that, for overall performance, $K_l=3$ and $K_f=512$ perform slightly better than other settings.
In the following experiments, we will use these two values as the default setting.

\begin{table}[]
\centering
\begin{tabular}{cccc}
\hline
$K_l$     & PPA & CAA & mIoU \\ \hline
0  & 85.9\%   & 54.2\%  & 38.9\%  \\
1  & 86.0\%   & 54.1\%  & 39.5\%  \\
2  & 86.6\%   & 52.7\%  & 38.7\%  \\
3  & 86.3\%   & 54.2\%  & 39.2\%  \\
4  & 85.5\%   & 53.6\%  & 38.0\%  \\ \hline
\end{tabular}
\caption{Results of using different $K_l$ for the contextual dependency predictor on SIFT Flow. $K_f$ is set to $128$.}
\label{tb_layer}
\end{table}

\begin{table}[]
\centering
\begin{tabular}{cccc}
\hline
 $K_f$    & PPA & CAA & mIoU \\ \hline
128  & 86.1\%   & 55.0\%  & 40.0\%  \\
256  & 86.9\%   & 54.5\%  & 40.2\%  \\
512 &  87.0\%   & 54.3\%  & 40.7\%  \\ \hline
\end{tabular}
\caption{Results of using different $K_f$ for the contextual dependency predictor on SIFT Flow. $K_l$ is set to $3$.}
\label{tb_chn}
\end{table}

\begin{table*}[]
\centering
\begin{tabular}{c|ccc|ccc}
\hline
               & \multicolumn{3}{c|}{PASCAL Context} & \multicolumn{3}{c}{COCO Stuff} \\ \hline
               & PPA        & CAA       & mIoU       & PPA      & CAA      & mIoU      \\
baseline-no    & 71.0\%          & 51.0\%        & 39.3\%         &  59.9\%       &      41.2\%    &   27.9\%        \\
baseline-ave   & 72.1\%          & 52.1\%        & 41.1\%         &    60.8\%      &      40.5\%    &    27.6\%       \\
ours           &    \textbf{72.8}\%        &   \textbf{54.4}\%        &    \textbf{42.0}\%        &    \textbf{61.6}\%      &   \textbf{42.5}\%       &   \textbf{29.1}\%        \\ \hline
\end{tabular}
\caption{Results of ablation study on PASCAL Context dataset and COCO Stuff dataset.}
\label{tb_ablation}
\end{table*}

\subsection{Ablation Study}

To investigate the usefulness and effectiveness of the SCA module,
we compare our method with two baselines that use different configurations of the SCA module.
Note that, for a fair comparison, all these baselines use the same convolution and deconvolution modules as our proposed network.
\begin{itemize}
\item \textbf{baseline-no} We remove the dependency predictor and set
$a_{ij}=1, \forall j = i$ and $a_{ij}=0, \forall j\neq i$.
In this case, the SCA module does not embed any context information into the local neuron representations, degenerating to a $1\times 1$ convolution layer.

\item \textbf{baseline-ave} We remove the dependency predictor and set $a_{ij}=1, \forall i,j\in[1,n]$.
In this case, the SCA module embeds the averaged context features into local neuron representations. This is similar to the global context module proposed in~\cite{Liu_15}, where the globally pooled feature is concatenated with the local feature everywhere.

%\item \textbf{baseline-share} We use the same convolution to extract both \emph{identity features} and \emph{context features}, \ie, we learn a single matrix and use it in place of $\mathbf{W}_c$ and $\mathbf{W}_d$ in \eqref{eq_context}.
%\zhwang{The training of this baseline on COCO Stuff can not be finished before deadline. I simply remove this one as it does not effect too much}

\end{itemize}

The results on both PASCAL Context and COCO Stuff are shown in Tab.~\ref{tb_ablation}.
As can be seen, baseline-no performs the worst on most metrics as it does
not explicitly embed context into local neuron representation. Although the use of dilated convolution can help to expand the theoretical receptive fields of  neurons to aggregate context, the empirical receptive field that affects the neurons may still be small~\cite{Liu_15}.
With the use of globally averaged features as context, the baseline-ave expand its receptive field to the entire image, thus improving the performance of baseline-no.
Our method further improves baseline-ave on all the metrics on both datasets
by a significant margin. Since all the settings are the same except for the way of aggregating context, it demonstrates the effectiveness of the proposed neuron-level selective context aggregation. The proposed SCA module successfully models neuron-dependent contextual information, by exploring the dependencies between neurons, to facilitate context-aware feature learning.

\begin{figure}[!htb]
\centering
\subfigure[]{
\includegraphics[width=0.48\textwidth]{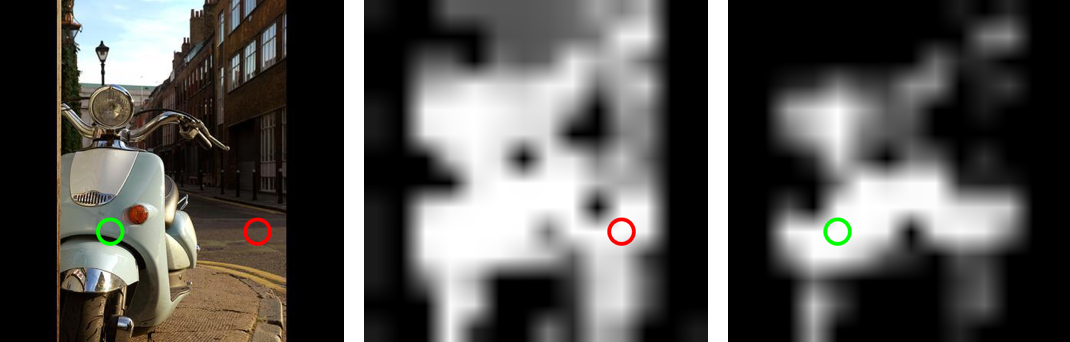}\label{fig_diff}}
\subfigure[]{
\includegraphics[width=0.48\textwidth]{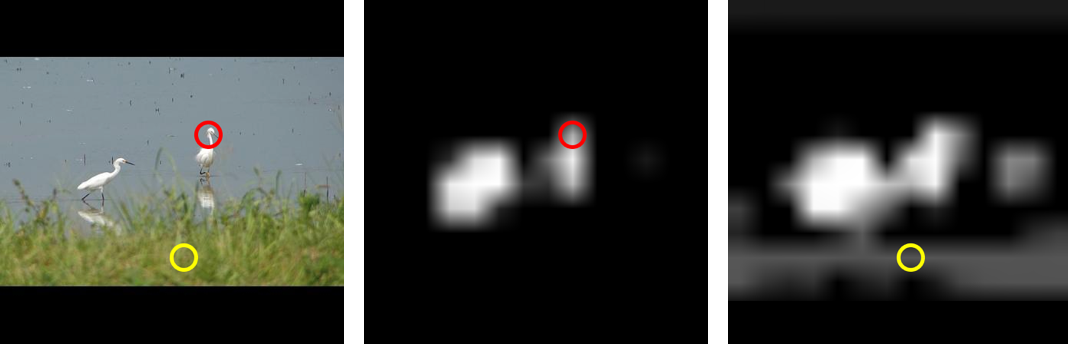}\label{fig_fixed}}
\caption{Visualization of dependency masks of different neurons. Red and green circles denote different regions corresponding to the selected neurons.}
\label{fig_mask}
\end{figure}
\subsection{Comparisons with the State of the Art}

\iffalse
\begin{table}[]
\centering
\caption{Comparison with state-of-the-art methods on SIFT Flow dataset.}
\label{tb_siftflow}
\begin{tabular}{llll}
\hline
Methods & PPA & CAA & mIoU \\ \hline
Byeon~\etal~\cite{Byeon_15}       &   70.1\%  &  22.6\%  & N/A \\
Liu~\etal~\cite{Liu_09}       &   74.8\%  &  N/A   & N/A\\
Farabet~\etal~\cite{Farabet_13}       &   78.5\%  &   29.4\% &  N/A\\
Pinheiro~\etal~\cite{Pinheiro_14}       &  77.7\%   &    29.8\%&  N/A\\
Tighe~\etal~\cite{Tighe_13}       &   79.2\%  &  39.2\%  &  N/A\\
Sharma~\etal~\cite{Sharma_14}       &  79.6\%   &  33.6\%  &  N/A\\
Yang~\etal~\cite{Yang_14}       &   79.8\%  &   48.7\%  & N/A \\ \hline
ParseNet~\etal~\cite{Liu_15}       &  86.8\%   &   52.0\% &  40.4\%\\
ConvPP-8s~\etal~\cite{Xie_16}       &  N/A &   N/A \% & 40.7\%\\
FCN-8s~\etal~\cite{Long_17}       &  85.9\%   &   53.9\% & 41.2\%\\
UoA-Context+CRF~\etal~\cite{Lin_16}       &  88.1\%   &   53.4\% &  44.9\% \\\hline
DAG-RNN~\etal~\cite{Shuai_17}       &   87.3\%  & 60.2\%   & 44.4\% \\\hline
ours          &  xxx\%   &   xxx\%  & xxx\% \\ \hline
\end{tabular}
\end{table}
\fi

\begin{figure*}[!htb]
\centering
\includegraphics[width=0.9\textwidth]{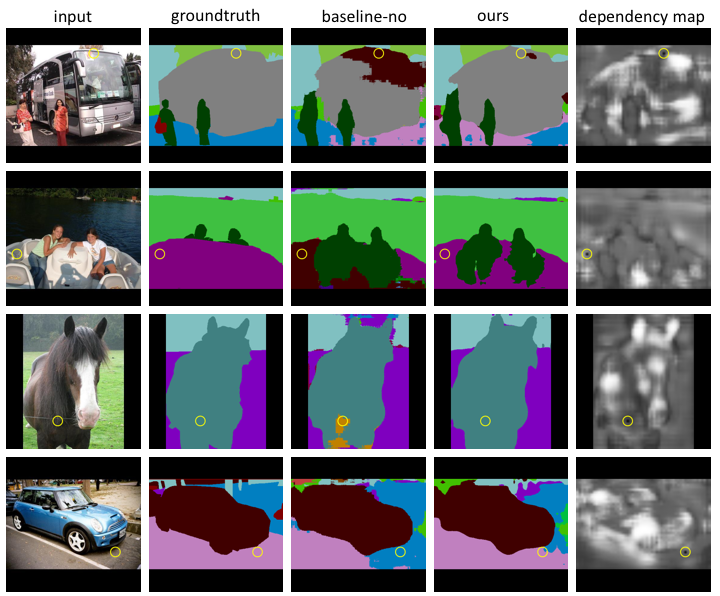}
\caption{Qualitative results selected in PASCAL Context dataset. We place a yellow circle on a selected region where most pixels inside are misclassified by the baseline-no method while are correctly classified by our method. The leftmost is the dependency mask corresponds to the select region.
}
\label{fig_pascal}
\end{figure*}

\begin{figure*}[!htb]
\centering
\includegraphics[width=0.9\textwidth]{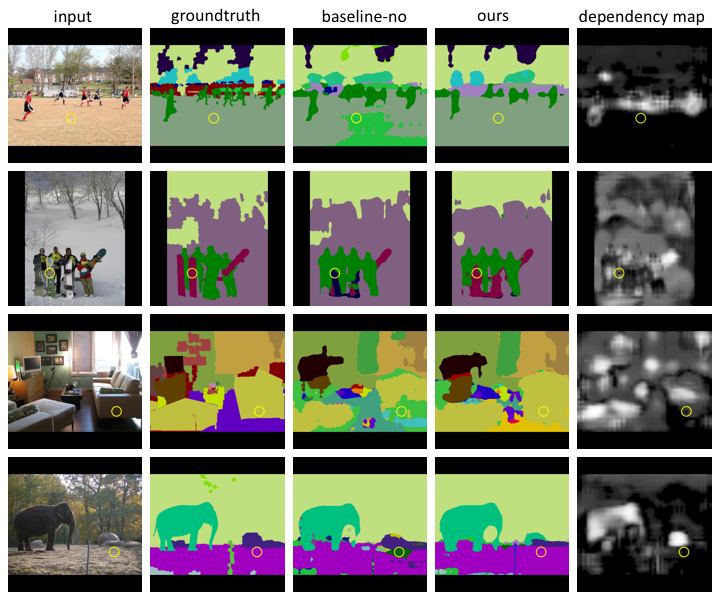}
\caption{Qualitative results selected in COCO Stuff dataset. We place a yellow circle on a selected region where most pixels inside are misclassified by the baseline-no method while are correctly classified by our method. The leftmost is the dependency mask corresponds to the select region.}
\label{fig_coco}
\end{figure*}

\begin{table}[htb]
\centering
\caption{Comparison with state-of-the-art methods on PASCAL Context dataset.}
\label{tb_pascal}
\begin{tabular}{llll}
\hline
Methods & PPA & CAA & mIoU \\ \hline
CFM~\etal~\cite{Dai_15}       &   N/A  &  N/A  & 31.5\% \\
DeepLab~\etal~\cite{Chen_16}       &   N/A  &  N/A   & 37.6\%\\
ParseNet~\etal~\cite{Liu_15}       &  N/A   &   N/A &  40.4\%\\
ConvPP-8s~\etal~\cite{Xie_16}       &  N/A &   N/A & 41.0\%\\
FCN-8s~\etal~\cite{Long_17}       &  67.5\%   &   52.3\% & 39.1\%\\
UoA-Context+CRF~\etal~\cite{Lin_16}       &  71.5\%   &   53.9\% &  \textbf{43.3}\% \\\hline
CRF-RNN~\etal~\cite{Zheng_15}       &  N/A   &    N/A &  39.3\%\\
DAG-RNN~\etal~\cite{Shuai_17}       &   72.7\%  & \textbf{55.3}\%   & 42.6\% \\\hline
ours          &  \textbf{72.8}\%   &   54.4\%  & 42.0\% \\  \hline
\end{tabular}
\end{table}

The results on PASCAL Content dataset presented in Tab.~\ref{tb_pascal}.
Comparing with previous CNN-based models,
our method significantly outperforms all these methods on all metrics except for mIoU on UoA-Context+CRF.
Considering RNN-based methods, our method outperforms CRF-RNN by a large margin. For DAG-RNN, we perform better on PPA, and are worse than it on CAA and mIoU.
The results on COCO Stuff are presented in Tab.~\ref{tb_coco}.
As can be seen, we also outperform all CNN-based methods, and perform better than DAG-RNN on CAA and worse than it on PPA and mIoU. The results on these two challenging datasets demonstrate the effectiveness of the proposed method.

We also note that in DAG-RNN, skip-connections are used to fuse the features of higher layers with low level convolution features, which significantly improves their performance.
In our method, to best investigate how the proposed SCA model helps to improve the representation ability of the last convolution features, we do not employ skip-connections in our architecture.
Meanwhile, the RNN-based methods have to sequentially update their hidden vectors over time, which make them inappropriate to be parallelized to improve speed. In contrast,
our method is a feed-forward network, which can be easily parallelized.

\begin{table}[htb]
\centering
\caption{Comparison with state-of-the-art methods on COCO Stuff dataset.}
\label{tb_coco}
\begin{tabular}{llll}
\hline
Methods & PPA & CAA & mIoU \\ \hline
FCN~\etal~\cite{Caesar_16}       &   52.0\%  &  34.4\%  & 22.7\% \\
DeepLab~\etal~\cite{Chen_16}   &   57.8\%  &  38.1\%  & 26.9\%\\
FCN-8s~\etal~\cite{Long_17}    &   60.4\%  &  38.5\%  & 27.2\%\\\hline
DAG-RNN~\etal~\cite{Shuai_17}  &   \textbf{62.2}\%  & 42.3\%   & \textbf{30.4}\% \\\hline
ours          &  61.6\%   &   \textbf{42.5}\%  & 29.1\% \\ \hline
\end{tabular}
\end{table}

\subsection{Qualitative Results}
Fig.~\ref{fig_mask} visualizes the dependency maps for different selected neurons, from which we note that there are  two different behaviors of the CDP.
Firstly, as shown in Fig.~\ref{fig_diff}, the two different selected neurons (indicated using green and red circles) generally have different dependency maps, as they belong to different classes. The CDP has learned to predict a different dependency map for each neuron, enhancing their local representation with different global context.
Secondly, as shown in Fig.~\ref{fig_fixed}, there are also cases where two different neurons may have very similar dependency maps.
This may be attributed to the fact that the salient regions, \eg, the birds here, provide an important context for all other regions. Still, note that context of the green neuron in the grass region includes the surrounding grass regions, while these regions are not important for the semantic interpretation of the red neuron containing the bird's head.

To better understand how the proposed SCA module helps to improve segmentation accuracy, we show some visual examples from Pascal \textit{Context} and COCO \textit{Stuff} in Fig.~\ref{fig_pascal} and Fig.~\ref{fig_coco}, respectively. These examples are also included in the supplementary material with the full legend of label colors. For example, in the top row of Fig.~\ref{fig_pascal} it may be seen that, without considering context, the top part of the bus is mislabeled as `car'. The dependency map for a neuron located in that region predicts high dependencies for other neurons in different parts of the bus, which are easier to recognize as such, and thus our method correctly labels the problematic area as `bus'. Similarly, in the top row of Fig.~\ref{fig_coco}, part of the playing field is mislabeled as `grass' in the baseline-no result. Neurons in that area are assigned a high dependency with the areas containing the players, which helps our method to correctly label the area as `playingfield'.

\section{Conclusion\label{sec_conclusion}}

In this paper, we have presented a neuron-level selective context aggregation module for scene segmentation.
The key idea is to augment local representation with global context which is selectively aggregated at each neuron according to its own dependencies, which are predicted on-the-fly by an auxiliary network.
The strength of our method stems from that fact that the auxiliary network is jointly trained with the main network without extra supervision. It learns to predict appropriated dependency coefficients for different pairs of neurons to selectively aggregate context by minimizing the overall segmentation loss.
This mechanism enables data-driven inference of contextual dependencies, and facilitates context-aware feature learning.
%\dc{The following might be like shooting in our own leg … it may be like %admitting that we didn't succeed. How about the following alternative below}?

%While recent RNN-based method achieved state-of-the-art performance on scene segmentation tasks, our method obtains comparable results to theirs.
%Moreover, unlike RNN-based method whose hidden vectors must be updated sequentially over time, our method is a feed-forward network which is easy to be parallelized to achieve high speed in run time.

Adding global context to local representation does not necessarily help in many cases. It mainly excels in harder cases, where otherwise local context alone might fail. In our current work, we learned general neuron dependencies. In the future, we consider narrowing down and focusing specifically on long-range dependencies, or possibly cross-image dependencies, where we analyze more than a single image in test time. We believe that learning dependency for co-segmentation is a viable research direction to further improve semantic segmentation capabilities.

{\small
\bibliographystyle{ieee}
\bibliography{egbib}
}

\end{document}